\documentclass[runningheads,a4paper]{llncs}
\usepackage[utf8]{inputenc}
\usepackage{amsmath}
\usepackage{amsfonts}
\usepackage{amssymb}
\usepackage{graphicx}
\usepackage{lscape}

\DeclareMathOperator{\I}{\mathcal{I}}

\title{Independence of Sources in Social Networks}

\author{Manel Chehibi\inst{1} \and Mouna Chebbah\inst{2} \and Arnaud Martin\inst{3}}
\institute{Univ. Manouba, ESEN, Tunisie\\ chehibimanel@gmail.com\and
LARODEC, Univ. Manouba, ESEN, Tunisie\\mouna.chebbah@esen.tn \and
IRISA, Universit\'e de Rennes1, Lannion, France\\Arnaud.Martin@univ-rennes1.fr
} 
\begin{document}

\maketitle

\paragraph{Abstract} Online social networks are more and more studied. The links between users of a social network are important and have to be well qualified in order to detect communities and find influencers for example. In this paper, we present an approach based on the theory of belief functions to estimate the degrees of cognitive independence between users in a social network. We experiment the proposed method on a large amount of data gathered from the Twitter social network.
\section{Introduction} 
Online social networks are online platforms that connect users. They have gained a lot of interest and popularity over the last decade. Many people rely on social networks particularly on information, news and opinions shared by users on diverse subjects.

An online social network, such as Twitter, helps users to share subjective information reflecting their personal opinions. In fact, in a social network, users become sources of information who produce different kinds of information (opinions, facts, news, rumors, etc.). However, some users are cognitively dependent on others. In addition, an online social network enable its users to interact with each other by several activities such as sharing, quoting, or commenting other users' posts. These users' interactions provide insights for the cognitive dependence/independence relationships among users in a social network. A user is supposed to be cognitively dependent on another user if he relies on and adopts information that he provides.

The aim of this paper is to study dependencies of sources in social networks. Information about sources' dependencies in a social network can be used to detect related groups, communities \cite{Kudelka11}, \ldots.\\
The identification of communities can help for targeted marketing. It can also be used for influence propagation \cite{Jendoubi17} to promote new products and define new marketing strategies. Indeed, a company wishing to launch a marketing campaign or a new product can use relations of dependencies to speed up the propagation.

In this paper, we propose an approach to estimate the degrees of independence/dependence  between users of a social network. Twitter is chosen as an example of a directed social network; thus, we detail the proposed measure using Twitter vocabulary.  The dependence relationship between users is an oriented relation; therefore, Twitter is very appropriate to illustrate our approach.\\
The proposed approach is based on the theory of belief functions to estimate uncertain degrees of independence between users. The theory of belief functions is used to asses uncertain degrees of belief on the independence of users. This theory is also chosen thanks to the great number of combination rules that merge subjective information.

The remainder of this paper is organized as follows: Section $2$ recalls some basic concepts of the theory of belief functions; Section $3$ details the proposed approach to estimate degrees of independence/dependence. Finally, Section $4$ presents an experimental study of our approach before concluding in section $5$.	
\section{Theory of belief functions}
The theory of belief functions, also called Dempster-Shafer theory, was first introduced by Dempster \cite{ref1} and mathematically formalized by Shafer \cite{ref2}. This theory models imprecise, uncertain and missing data.

In the theory of belief functions, a \textit{frame of discernment}, noted \linebreak $\Theta=\lbrace H_1,...,H_N \rbrace$, is a set of $N$ exhaustive and mutually exclusive hypotheses $H_i, 1 \le i \le N$. only one of them is likely to be true.

The \textit{power set}, $2^\Theta=\lbrace A / A \subseteq \Theta \rbrace=\lbrace \emptyset,H_1,...,H_N,H_1\cup H_2,...,\Theta \rbrace$, enumerates $2^N$ sub-assemblies of $\Theta$. It includes not only hypotheses of $\Theta$, but also, disjunctions of these hypotheses.

The true hypothesis in $\Theta$ is unknown; thus, a degree of belief is assessed to subsets of $2^\Theta$ reflecting our degree of faith on the truth of each subset of $2^\Theta$.

A \textit{basic belief assignment (bba)}, also called \textit{mass function}, is noted $m^\Theta$ and defined such that:
\begin{equation}
\begin{split}
& m^\Theta : 2^\Theta \rightarrow [0,1]\\
& m^\Theta(\emptyset)=0\\
& \sum\limits_{A \subseteq \Theta}m(A)=1\\
\end{split}
\end{equation}
The mass $m^\Theta(A)$ represents the degree of belief on the truth of $A \in 2^{\Theta}$. When $m^\Theta(A)>0$, $A$ is called \textit{focal element}.

In the theory of belief functions, decision is generally made using \textit{pignistic probabilities} \cite{ref3}.  
The pignistic probability, noted $BetP^\Theta$, is deduced from $m^\Theta$ as follows:
\begin{equation}
BetP(H_i) = \sum\limits_{\substack{A \in 2^\Theta \\ H_i \subset A}}\frac{1}{\mid A\mid}m^\Theta(A) \qquad  \forall  H_i  \in  \Theta
\end{equation}
where $\mid A \mid$ is the number of hypotheses which train it.

In the theory of belief functions, combination rules are proposed to merge distinct mass functions in order to produce a more reliable information. It consists on building an unique mass function by combining several elementary mass functions arising from multiple distinct sources of information. 

\paragraph{Dempster's rule of combination} \cite{ref1} is the first rule that merges several mass functions provided by distinct and independent sources. The combination of two mass functions $m^\Theta_{S_1}$ and $m^\Theta_{S_2}$ provided by $S_1$ and $S_2$ is given as follows:
\begin{equation}
\label{Eqoplus}
m^\Theta _{1\oplus 2}(A)=(m^\Theta _1 \oplus m^\Theta _2)(A)=\left\{
\begin{tabular}{ll}
$\frac{\displaystyle{\sum _{B\cap C=A}} m^\Theta _1(B)\times m^\Theta _2(C)}{1-\displaystyle{\sum _{B\cap C=\emptyset}} m^\Theta _1(B)\times m^\Theta_2(C)}$&$\forall A\subseteq\Theta,\hspace{0.1cm}A\neq \emptyset$\\
$0$&$\mbox{if}\hspace{0.1cm}A=\emptyset$\\
\end{tabular}
\right.
\end{equation}

The reliability of an evidential information is not always insured. In fact, an evidential data can be supplied by a partially reliable or an unreliable source. In order to take the source's reliability into account, its beliefs are discounted proportionally to its reliability. Let $\alpha\in\left[ 0,1\right]$ be the reliability of a source $ S_1 $ and $m^\Theta$ a mass function provided by $S_1$. The \textit{discounting} of $m^\Theta$ produces $^\alpha m^\Theta$ defined by: 
\begin{equation} 
\begin{cases}
^\alpha m^\Theta(A)= \alpha \times m^\Theta(A) & \text{if },\forall A  \subset  \Theta\\
^\alpha m^\Theta(\Theta)= 1- \alpha \times (1-m^\Theta( \Theta ))\\ 
\end{cases} 
\end{equation}
\section{Uncertain Measure of Independence in Twitter}
Many researches are focused on measuring the independence in several social networks. 
Leenders \cite{Leeders02} proposed and approach focused on the opinions and attitudes of users in a social system. These opinions and attitudes are shaped by social influence. The proposed approach depend partially on individual characteristics.

 Kudelka et al. \cite{Kudelka11} makes use of the measurement of dependence between the network vertices for the detection of communities in social networks. 

To predict a user actions (behaviors) in a social network, Tan et al. \cite{Tan10} consider diverse factors: the influence from his friends, the \textit{correlation} between users' actions and his historic behaviors. They conducted an experiment on Twitter and they found that more friends perform the action, a user also tends to perform the action and the likelihood that two friends perform an action at the same time is always larger than the likelihood that randomly two users perform the same action at the same time.

Jendoubi et al. \cite{Jendoubi17} propose to detect influencer in Twitter using the theory of belief functions. They consider three Twitter metrics to quantify the influence between users: followers, mention, retweet. 

Twitter is a social network that enables its users to establish many types of relation between them. A relation between users of Twitter may be a \textit{follow}, a \textit{retweet}, a \textit{mention} or a \textit{citation}.

 These ties are considered as dependence indexes for the several reasons: First, the retweet actions represent the amount of information tweeted by a user from the tweets of another user. This amount reflects the degree of adoption of the opinions of other users. \\
Then, the mention represents the quantity of messages directly sent to other specific users in order to establish direct communications with them. These actions reflect the importance of a part of the Twitter users and their ideas for other users in the network.\\
Finally, the citation represents the degree of reliance of some users on other users by citing them in their tweets.

Therefore, we consider that  degrees of dependence between users of Twitter can be deduced from numbers of follows, retweets, mentions and citations. In this paper, we propose to estimate degrees of cognitive dependence between users of Twitter.  Two users are cognitively dependent when information provided by a user are affected by the information produced by the other one. We note that the cognitive independence is matter of researches in the theory of belief functions \cite{ChebbahMY15}.  Two variables \cite{ref2} are assumed to be cognitively independent with respect to a belief function if any new evidence that appears on only one of them does not change the evidence of the other variable. In addition, two sources \cite{ChebbahMY15} are cognitively independent if they do not communicate and if their evidential corpora are different. Two sources are either positively are negatively dependent; in the case of negative dependence, sources are dependent but their ideas are different. Otherwise, influencers \cite{Jendoubi17} are sources that have a maximum of impact in the ideas of others. Dependence and influence measures are different but quite similar. Thus, the dependence measure may be used for influence maximization. 

A user $u$  in Twitter is cognitively dependent on another user $v$ if $u$ is following $v$ and $u$ frequently retweets tweets of $v$ or/and, $u$ frequently mentions $v$ in his tweets.

\begin{figure}[!h]
\label{fig1}
\center
\includegraphics[scale=0.5, trim=0 0.5cm 0 2cm]{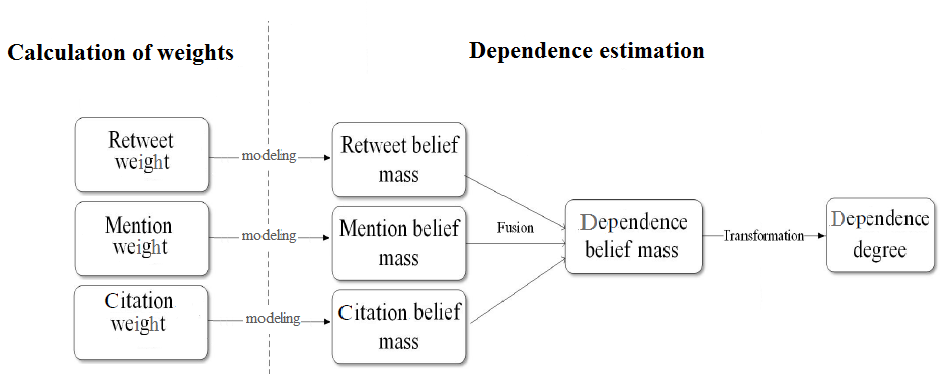}
\caption{The general framework of the proposed approach} 
\label{swap}
\end{figure}

 Figure \ref{fig1} shows the proposed approach to estimate independence of users in Twitter. The proposed approach is in $2$ steps:
\begin{enumerate}
\item In the first step, weights are estimated. Thus, we define a weight for each aspect of dependence: retweet, mention and citation.
\item In the second step, the independence estimation. In this step, we use the theory of belief functions to (i) model each independence aspect, (ii) to combine them and (iii) to make a decision regarding the independence of users.
\end{enumerate}
\subsection{Step $1$: Estimation of weights}
In Twitter, a user $u$ following a user $v$  can retweet, mention or/and cite $v$. Each information about the retweet, mention or/and citation may reflect the dependence or the independence of $u$ on $v$. Thus, a vector of weights $(w_r,w_m,w_c)$ is assigned to each link $(u,v)$ as shown in figure \ref{swap}. Note that $u$ is following $v$ and the vector of weights will be used to learn the independence/dependence of $u$ on $v$.
\begin{figure}[!h]
\center
\includegraphics[width=4cm,height=2cm]{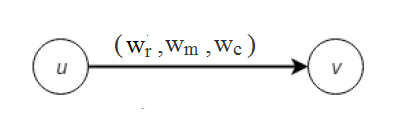}
\caption{Weight vector between u and v} 
\label{swap}
\end{figure}
Let $G = (V,E)$ be the social network where $V$ is the set of nodes, $E$ is the set of links, $u \in V$ is a follower of $v \in V$ in Twitter.
The weights $w_r$, $w_m$ and $w_c$ of the link $(u,v) \in E$ are estimated using the following measures:
\begin{enumerate}
\item The retweet weight, $w_r(u,v)=\dfrac{Rt_u(v)}{Rt_u}$, is the weight defining the number of times that $u$ has retweeted the tweets of $v$; $ Rt_u(v) $ is the number of tweets of $v$ that were retweeted by $u$ and $ Rt_u $ is the total number of retweets of $u$.\\
\item The mention weight, $w_m(u,v)=\dfrac{Mt_u(v)}{Mt_u}$, is the weight defining the number of times that $u$ mentioned $v$ in his tweets; $ Mt_u(v) $ is the number of tweets of $u$ in which $v$ was mentioned and $ Mt_u $ is the total number of mentions of $u$.\\
\item The citation weight, $ w_c(u,v)=\dfrac{Ct_u(v)}{Ct_u} $, is the weight defining the number of times that $u$ quoted the tweets of $v$; $ Ct_u(v) $ is the number of tweets of $v$ who have been quoted by $u$ and $ Ct_u $ is the total number of citations of $u$.
\end{enumerate}
\subsection{Step $2$: Independence estimation}
The dependence estimation is based on the defined weights. Let $G=(V,E,W)$ be a directed graph where $W$ is the set of weights' vectors, such that \linebreak $(w_r(u,v), w_m(u,v),$ $w_c(u,v)) \in W$ is the weight vector associated to the link $(u,v)$. The independence estimation process is in three basic steps:
\begin{enumerate}
\item In the first step, a mass function is built from each weight on the link. Let $ \I = \lbrace D, I \rbrace $ be the frame of discernment of the independence where $D$ is the hypothesis that users are dependent and $I$ is the hypothesis that users are independent. Mass functions are estimated as follows:
\begin{enumerate}
\item First, the retweet weight justifies our belief on the independence of users. Therefore, $m_{r_{(u,v)}}^{\I}$ is defined as follows:
\begin{equation}
\left\{
\begin{tabular}{l}
$m_{r_{(u, v)}}^{\I} (D) = \alpha_{r_u} \times w_r(u, v)$\\
$m_{r_{(u, v)}}^{\I}(I) = \alpha_{r_u} \times (1-w_r(u, v))$\\
$m_{r_{(u, v)}}^{\I}(I,D) = 1-\alpha_{r_u}$\\
\end{tabular}
\right.
\end{equation}

Note that $\alpha_{r_u}=\dfrac{Rt_u}{T_u}$ is a discounting coefficient that takes into account the total number of tweets $ T_u $. The mass function $m_{r_{(u, v)}}^{\I}$ is more reliable when the number of retweets is enough big in comparison with the total number of tweets. 
For example, assume that a user $u$ has posted twenty eight tweets in two weeks and that among these tweets there are ten retweets, seven of them are from $v$. Without discounting using $ \alpha_{r_u} $, the value of $m_{r_{(u, v)}}^{\I}(D)$ will be equal to $ 0.7 $ which does not reflect the reality. In fact, the number of tweets that $u$ has retweeted $v$ represents only the quarter of the total number of tweets of $u$. 

\item Then, a mass function $m_{m_{(u, v)}}^{\I}$ is deduced from the mention weight as follows:
\begin{equation}
\left\{
\begin{tabular}{l}
$m_{m_{(u, v)}}^{\I} (D) = \alpha_{m_u} \times w_m(u, v)$\\
$m_{m_{(u, v)}}^{\I} (I) = \alpha_{m_u} \times (1-w_m(u, v))$\\
$m_{m_{(u, v)}}^{\I} (I,D) = 1-\alpha_{m_u}$\\
\end{tabular}
\right.
\end{equation}
Where $\alpha_{m_u}=\dfrac{Mt_u}{T_u}$ is a discounting coefficient. The discounting coefficient $ \alpha_{m_u}$ is used to take into account the total number of tweets quoted by $u$ with respect to the total number of tweets of $u$.

\item Finally, the mass function $m_{c_{(u, v)}}^{\I}$ is deduced from the citation weight as follows:
\begin{equation}
\left\{
\begin{tabular}{l}
$m_{c_{(u, v)}}^{\I} (D) = \alpha_{x_u} \times w_c(u, v)$\\
$m_{c_{(u, v)}}^{\I}(I) = \alpha_{c_u} \times (1-w_c(u, v))$\\
$m_{c_{(u, v)}}^{\I} (I,D) = 1-\alpha_{c_u}$\\
\end{tabular}
\right.
\end{equation}
Where $\alpha_{c_u}=\dfrac{Ct_u}{T_u}$ is a discounting coefficient that takes into account the total number of tweets of $u$ mentioning $v$ with respect to the total number of tweets of $u$.
\end{enumerate}

\item Then, mass functions $m_{r_{(u, v)}}^{\I}$, $m_{m_{(u, v)}}^{\I} (D)$ and $m_{c_{(u, v)}}^{\I}$ are combined with Dempster's rule of combination as follows: \\
\begin{equation}
m^{\I}_{(u,v)}= m_{r_{(u, v)}}^{\I} \oplus m_{m_{(u, v)}}^{\I}\oplus m_{c_{(u, v)}}^{\I}
\end{equation}

\item Finally, degrees of independence $Ind(u,v)$  and dependence $Dep(u,v)$ corresponds to pignistic probabilities computed from the combined mass function $m^{\I}_{(u,v)}$ as follows:
\begin{equation} 
\begin{cases}
Dep(u,v)= BetP(D)\\
Ind(u,v)= BetP(I)
\end{cases} 
\end{equation}
We have:
\begin{equation}
 Dep(u,v) +Ind(u,v)= 1
\end{equation}

The dependence degree $Dep(u,v)$ is non-negative, it is either positive or null. It is also normalized. In fact, the degree of dependence $Dep(u,v)$ is a degree that lies in the interval $[0,1]$.
When $Dep(u,v)=1$, $u$ is totally dependent on $v$ ; $Dep(u,v)=0$ implies that $u$ is totally independent of $v$.  
Decision is made according to the maximum of pignistic probabilities. If $Dep(u,v)\geq Ind(u,v)$ then $u$ is dependent on $v$, in the opposite case, if $Ind(u,v)>Dep(u,v)$, $u$ is independent from $v$.
\end{enumerate}
\section{Experiments}
The proposed approach is tested on data collected from Twitter; because it is a directed social network that provides a large number of messages published per day. Unlike other social media platforms like Facebook, the content of Twitter is public and accessible via programming interfaces. In our experimental study, we used the Twitter streaming API through a Python library called Tweepy. This library provides access to Twitter data {\em via} its programming interface, Twitter API. The Twitter Streaming API allows retrieving data in real-time. It allows also filtering tweets by several keywords or according to their geographical position. In our case, we are interested in collecting tweets written by specific users. For this purpose, we filtered tweets by a list of users IDs. We crawled Twitter data for the period between 05/06/2017 and 13/8/2017. We get an important number of tweets (205271 tweets) corresponding to 10350 users on this period. Experiments of the proposed approach detailed in this section are made on a large number on users, tweets, retweets, mention and citation as detailed in table \ref{data}. Note that retweets, mentions and citations are considered as tweets.
\begin{table}[!h]\footnotesize
\caption{Data Collected from 05/06/2017 to 13/8/2017}
\label{data}
\begin{center}
\begin{tabular}{|c||c||c||c||c|}
\hline
 Users &  Tweets &  Retweets & Mentions&  Citations\\
\hline
$10350$&$205271$&$32842$&$71901$&$14613$ \\
\hline
\end{tabular}
\end{center}
\end{table}
\begin{itemize}

\item Table \ref{expD}, shows that there are independent relationship between a part of users despite there are a follow relationship between them. For example, the user $S_1$ is independent from the user $S_2$ and the same for the user $S_3$ with $S_{29}$ with a lower degree of dependence. All experiments are made on real data described on table \ref{data} which are collected from Tweeter. Users are numbered to respect the anonymity and privacy. Therefore, the follow relationship in Twitter does not necessarily imply the cognitive dependence between users.  
In an explicit way, a user $u$ who follows another user $v$ in Twitter can be either cognitively independent or dependent on $v$.
\begin{table}[!h]\footnotesize
\caption{Examples of independence relationship}
\label{expD}
\begin{center}
\begin{tabular}{|c||c|}
\hline
Link & The degree of dependence\\
\hline
$ \left( S_{1},S_{2} \right) $ & $0.1$ \\
\hline
$ \left( S_{3},S_{29} \right) $ & $0.3$\\
\hline
$ \left( S_{4},S_{37} \right) $ & $0.2$\\
\hline
\end{tabular}
\end{center}
\end{table}
 
\item Table \ref{ass} shows that in the case where a user $u$ is dependent on a user $v$, $v$ is not necessarily dependent on $u$. In the case where a user $u$ is independent on a user $v$, $v$ is not necessarily independent on $u$.
\begin{table}[!h]\footnotesize
\caption{Examples of asymmetrical relationships}
\label{ass}
\begin{center}
\begin{tabular}{|c||c|}
\hline
Link & The degree of dependence \\
\hline
\hline
$\left( S_{8},S_{35} \right)$ &$0.6$\\
\hline
$\left( S_{35},S_{8} \right)$ &$0.2$\\
\hline
\hline
$\left( S_{10},S_{13} \right)$ &$0.7$\\
\hline
$\left( S_{13},S_{10} \right)$ &$0.3$\\
\hline
\end{tabular}
\end{center}
\end{table} 
\item Table \ref{depInd} shows that if users $u$ and $v$ are mutually independent or dependent, degrees of independence or dependence are not necessarily equal.

\begin{table}[!h]\footnotesize
\caption{Examples of mutual independence/dependence  with different degrees of independence/dependence}
\label{depInd}
\begin{center}
\begin{tabular}{|c||c|}
\hline
Link & The degree of dependence \\
\hline
\hline
$\left( S_{11},S_{5} \right)$ & $0.7$ \\
\hline
$\left( S_{5},S_{11} \right)$ & $0.6$\\
\hline
\hline
$\left( S_{12},S_{23} \right)$ & $0.3$ \\
\hline
$\left( S_{23},S_{12} \right)$ & $0.1$ \\
\hline
\end{tabular}
\end{center}
\end{table}
\end{itemize}

Tests are made on data collected from 05/06/2017 to 13/08/2018 as detailed in table \ref{data}. Degrees of independence and dependence are computed of each pair of users from the $10350$. Thus, degrees of independence and dependence are computed for each couple of users $(u,v)$ for all the $10350$ users. Note that for each couple of users we compute $Ind(u,v)$ and $Ind(v,u)$. Therefor $10350!*2$ values of independence are computed. In the complete graph there are $10350$ nodes, each node represents a user and $2$ values of independence for each couple of users. For tests, we have also estimated the degree of independence/dependence for users without any relationship of follow.\\

\begin{figure}[!h]
\includegraphics[trim= 5.5cm 0cm 0cm 0cm,scale=0.75]{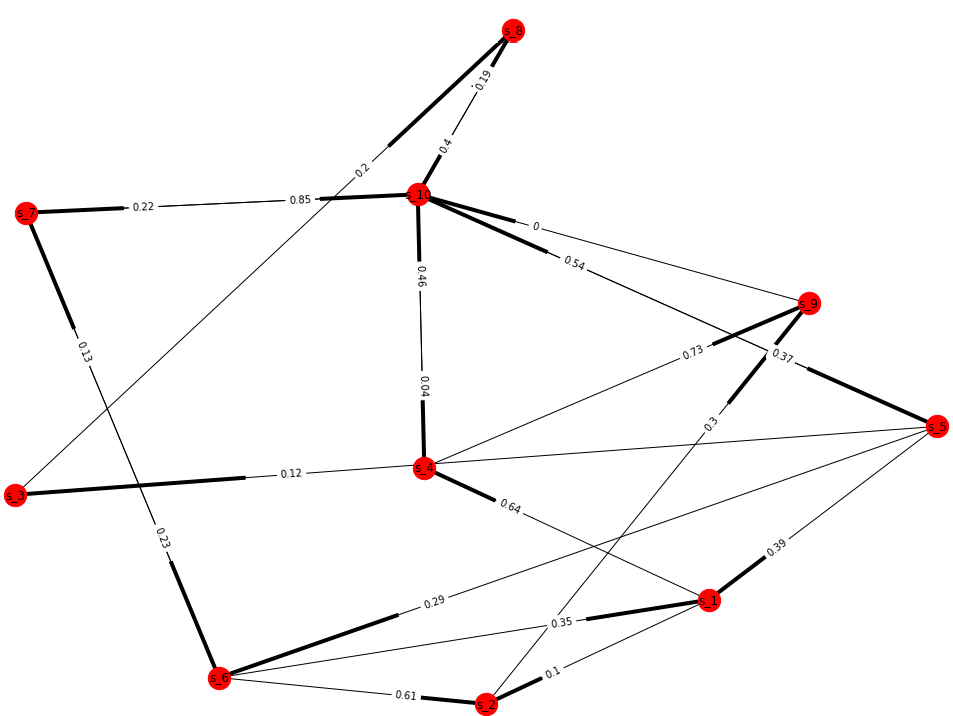} 
\caption{Example dependence analysis between users.} 
\label{graph}
\end{figure}

The dependence graph of figure \ref{graph} is a part of the complete graph. In figure \ref{graph}, only $10$ users from the $10350$ users are represented. These $10$ users are randomly chosen for simplicity seek and also to have a readable graph. Black links represent a follow link, the bold part links reflects the direction of follows. In other words, $S_1$ is following $S_2$ and $S_4$; $S_2$ is following $S_9$; $S_3$ is following $S_8$; $S_4$ is following $S_3$, $S_{10}$ and $S_9$; $S_5$ is following $S_6$, $S_{10}$ and $S_1$; $S_6$ is following $S_1$, $S_2$ and $S_7$; $S_7$ is following $S_{10}$ and $S_6$; $S_8$ is following $S_{10}$; $S_9$ is following $S_{10}$ and finally $S_{10}$ is following $S_4$, $S_5$, $S_7$ and $S_8$. Note that $(S_4,S_{10})$, $(S_5,S_{10})$, $(S_6,S_7)$, $(S_7, S_{10})$, $(S_8,S_{10})$ are mutually following each other.\\

Figure \ref{graph} shows that some users are cognitively dependent, for example $S_1$ is dependent on $S_4$ with a degree $0.64$; $S_4$ is dependent on $S_9$ and $Dep(S_4,S_9)=0.73$; $S_5$ is dependent on $S_{10}$ and $Dep(S_5,S_{10})=0.54$; $S_6$ is dependent on $S_2$ and $Dep(S_6,S_2)=0.61$; finally $S_7$ is dependent on S10 and $Dep(S_7,S_{10})=0.85$.\\

Finally, $(S_1,S_2)$, $(S_2,S_9)$, $(S_3,S_8)$, $(S_4,S_3)$, $(S_4,S_{10})$, $(S_5,S_6)$, $(S_5,S_1)$, $(S_6,S_1)$, $(S_6,S_7)$, $(S_7,S_6)$, $(S_8,S_{10})$, $(S_9,S_{10})$, $(S_{10},S_4)$,$(S_{10},S_5)$, $(S_{10},S_7)$ and $(S_{10},S_{8})$ are independent. Note that $S_{10}$ and $S_4$, $S_{10}$ and $S_8$, $S_7$ and $S_6$ are mutually independent.


\begin{table}[!h]\footnotesize
\caption{Dependence between users without any link of follow}
\label{depus}
\begin{center}
\begin{tabular}{|c||c|}
\hline
Users & The degree of dependence \\
\hline
$S_1$, $S_4$&$Dep(S_4,S_1)=0.21$\\
\hline
$S_1$, $S_3$&$Dep(S_1,S_3)=0.08$\\
&$Dep(S_3,S_1)=0.16$\\
\hline
$S_2$, $S_4$ & $Dep(S_2,S_4)=0.19$ \\
 & $Dep(S_4,S_2)=0.25$ \\
\hline
$S_5$, $S_6$ & $Dep(S_6,S_5)=0.15$\\
\hline
\hline
$S_3$, $S_8$ & $Dep(S_8,S_3)=0.13$ \\
\hline
$S_3$, $S_7$ & $Dep(S_3,S_7)=0.1$ \\
& $Dep(S_7,S_3)=0.21$ \\
\hline
\end{tabular}
\end{center}
\end{table}

Table \ref{depus} shows that users without any follow are independent. For example there is no follow between $S_1$ and $S_3$ because $S_1$ is not following $S_3$ and $S_3$ is not following $S_1$. Users $S_1$ and $S_3$ are mutually independent.
Users that are not following others are independent. When a user $u$ is not following another user $v$, $u$ is necessarily independent from $v$.  

\section{Conclusion}

Studying cognitive independence relationship among the Twitter social network users is a very important research topic since this online social network is widely used to post and share information. 
In fact, quantify the degrees of dependence between users can be very useful to disseminate information to the largest number of users which is a very important thing in many fields such as marketing.

Most of existing works that try to study the dependence between users in a social network, use only the network structure to measure the dependence of a user on another user and ignore many interesting dependence aspects. 
Nevertheless, the dependence measures that is based only on the network structure is not adequate to quantify the dependence between sources. In fact, in the twitter social network, a user can follow another user in the network without being necessarily cognitively dependent on him.

In this work, we propose an approach based on the theory of belief functions for measuring the dependence degrees between users in Twitter. We consider three dependence aspects witch are the retweets, the mentions and the citations and we use the Dempster-Shafer theory to model each dependence aspect, to combine them with taking into consideration the conflict that can arise between them and to make a decision with regard to the dependence a user
on another user in the network.

The results of the experimental study of our proposed approach show that the follow relationship in twitter does not necessarily imply the cognitive dependence between users and that the more the number of retweets, citations or/and mentions increase, the more the degree of dependence of a user on an other user increases and vice versa. It shows also that the dependence relationship between two users is not necessarily mutual and the dependence degrees between them are not necessarily equal.

As a future work, we will use our approach to detect communities in social networks.

\bibliographystyle{unsrt}
\bibliography{References}

\end{document}